\documentclass[sigconf]{acmart}
\AtBeginDocument{%
  }

\usepackage{graphicx}
\usepackage{subcaption}
\usepackage{multirow}
\usepackage{balance} 

\copyrightyear{2025}
\acmYear{2025}
\setcopyright{acmlicensed}\acmConference[CIKM '25]{Proceedings of the 34th ACM International Conference on Information and Knowledge Management}{November 10--14, 2025}{Seoul, Republic of Korea}
\acmBooktitle{Proceedings of the 34th ACM International Conference on Information and Knowledge Management (CIKM '25), November 10--14, 2025, Seoul, Republic of Korea}
\acmDOI{10.1145/3746252.3761348}
\acmISBN{979-8-4007-2040-6/2025/11}
\begin{document}

\title{WDformer: A Wavelet-based Differential Transformer Model for Time Series Forecasting}

% \author{Xiaojian Wang}
% \affiliation{%

\author{Xiaojian Wang}
\orcid{0009-0005-3042-5477}
\affiliation{%
  \institution{School of Computer Science and Technology, Zhejiang Normal University}
  \city{Jinhua}
  \country{China}}
\email{xiaojian@zjnu.edu.cn}

\author{Chaoli Zhang}
\authornote{Corresponding author.}
\orcid{0000-0003-4059-8396}
\affiliation{%
  \institution{School of Computer Science and Technology, Zhejiang Key Laboratory of Intelligent Education Technology and Application, Zhejiang Normal University}
  \city{Jinhua}
  \country{China}
}
\email{chaolizcl@zjnu.edu.cn}

\author{Zhonglong Zheng}
\orcid{0000-0002-5271-9215}
\affiliation{%
  \institution{School of Computer Science and Technology, Zhejiang Key Laboratory of Intelligent Education Technology and Application, Zhejiang Normal University}
  \city{Jinhua}
  \country{China}
}
\email{zhonglong@zjnu.edu.cn}

\author{Yunliang Jiang}
\orcid{0000-0003-4500-5836}
\affiliation{%
  \institution{Zhejiang Key Laboratory of Intelligent Education Technology and Application, School of Computer Science and Technology, Zhejiang Normal University}
  \city{Jinhua}
  \country{China}
}
\email{jyl2022@zjnu.cn}

\thanks{
This work was supported in part by the NSFC under Grant No. 62502456; in part by the Zhejiang Provincial Natural Science Foundation of China under Grant No. LQN25F020020;
in part by the Zhejiang Jinhua Science and Technology Project under Grant No. 2024-4-019; in part by the NSFC under Grant No. 62272419 and in part by Open Research Fund of Zhejiang Key Laboratory of Intelligent Education Technology and Application under Grant No. 2025ZNJYKF013.
}
% \affiliation{%

% \affiliation{%

% \affiliation{%

% \affiliation{%

% \renewcommand{\shortauthors}{Wang et al.}

\begin{abstract}
  Time series forecasting has various applications, such as meteorological rainfall prediction, traffic flow analysis, financial forecasting, and operational load monitoring for various systems. Due to the sparsity of time series data, relying solely on time-domain or frequency-domain modeling limits the model's ability to fully leverage multi-domain information. Moreover, when applied to time series forecasting tasks, traditional attention mechanisms tend to over-focus on irrelevant historical information, which may introduce noise into the prediction process, leading to biased results. We proposed WDformer, a wavelet-based differential Transformer model. This study employs the wavelet transform to conduct a multi-resolution analysis of time series data. By leveraging the advantages of joint representation in the time-frequency domain, it accurately extracts the key information components that reflect the essential characteristics of the data. Furthermore, we apply attention mechanisms on inverted dimensions, allowing the attention mechanism to capture relationships between multiple variables. When performing attention calculations, we introduced the differential attention mechanism, which computes the attention score by taking the difference between two separate $\mathrm{softmax}$ attention matrices. This approach enables the model to focus more on important information and reduce noise. WDformer has achieved state-of-the-art (SOTA) results on multiple challenging real-world datasets, demonstrating its accuracy and effectiveness. Code is available at~\href{https://github.com/xiaowangbc/WDformer}{https://github.com/xiaowangbc/WDformer}.
\end{abstract}





\maketitle
\section{Introduction}
Time series forecasting holds profound practical significance and extensive applicability in real-world contexts. It provides crucial decision support 
for various fields such as finance, transportation, education, healthcare, energy, and many others. Multiple deep learning models have achieved significant success in time series forecasting tasks~\cite{jia2024witran,luo2024moderntcn,wu2022timesnet,cai2024msgnet}. The Transformer model, due to its outstanding performance in the fields of Natural Language Processing (NLP)~\cite{xiong2020layer,devlin2018bert} and Computer Vision (CV)~\cite{dosovitskiy2020image,liu2021swin}, has received a great deal of attention from researchers. Researchers actively explore the application of Transformer models in time series forecasting tasks. Transformer-based forecasting models, leveraging the powerful long-range dependency-capturing capability endowed by the attention mechanism, have achieved remarkable success in time series forecasting tasks~\cite{zhou2021informer,wu2021autoformer}. The iTransformer~\cite{liu2023itransformer} treats independent time series as tokens and uses a self-attention mechanism to capture the multivariate correlations among them. Despite its simplicity in design, it has demonstrated remarkable effectiveness.

%However, due to the sparsity of time series data, finding correlations between multiple variables in time series is challenging.\@  % 这句跟后面的没啥联系，建议去掉。
Time series data can be analyzed from both the time domain and the frequency domain, with both aspects jointly forming the important informational basis for time series modeling. In the time domain, we focus on the arrangement of data points along the time axis. This allows us to discern trends, seasonal patterns, and other temporal characteristics within the data. In the frequency domain, our primary focus is on the frequency components that constitute the time series. This approach enables us to quickly identify periodic patterns that are not readily apparent in the time domain. By filtering frequencies, we can also effectively remove noise, thereby gaining a clearer understanding of the essential features of the data. When the attention mechanism is confined to the time domain for multivariate correlation modeling, the lack of information on frequency domain features often constrains the predictive performance of the model.\@ The Fourier transform can convert time series data from the time domain to the frequency domain, thereby revealing their frequency characteristics. However, this process results in the loss of time-domain information, making it impossible to fully utilize the details and structures of the time series in the time domain. As a method of multiresolution analysis, the wavelet transform is capable of decomposing time series data into wavelet functions of different scales and positions, thereby providing detailed information about the time series in both the time domain and the frequency domain simultaneously.  Therefore, we perform a wavelet transform on the time series to extract multi-scale information of the time series. Specifically, we apply the wavelet transform to the original time series data to obtain wavelet coefficients. For each wavelet coefficient, we perform embedding separately and then concatenate them. As a result, we obtain a matrix that contains both time and frequency information.

To reduce noise, we also incorporate a differential attention mechanism~\cite{ye2024differential} into our model. The differential attention mechanism was initially applied to NLP tasks, highlighting that the calculation of $\mathrm{softmax}$ attention matrices tends to focus excessively on irrelevant information. The differential attention mechanism takes the difference between two independently computed $\mathrm{softmax}$ attention matrices as the final $\mathrm{softmax}$ attention matrix. Natural Language Processing tasks and time series tasks show significant similarities in many aspects. From the perspective of data structure, both are based on sequential data and share certain similarities in form. What's more critical is that they both contain strong sequential dependencies, that is, the meaning or value of each element in the sequence is often closely linked to its adjacent preceding and succeeding elements. This dependency relationship is vital for a deep understanding and accurate processing of these tasks.\@ 
The core objective of the differential attention mechanism is to effectively reduce “attention noise” while significantly enhancing the model’s focus and capture ability for key information. It is worth noting that, compared to the common types of noise in the NLP field (such as spelling errors), the noise contained in time series data (such as sensor errors, abnormal values) is not only occurs more frequently, but also tends to be more concealed and complex in its manifestations.
The differential attention mechanism is also promising in reducing noise in time series forecasting tasks, so we incorporate it into our model. In experiments, our model achieved more accurate results on several real-world forecasting benchmarks than most of the existing models.

In this paper, our contributions are as follows:
\begin{itemize}
    \item We apply the wavelet transform to the original time series to obtain multi-level wavelet coefficients, and then embedding each level of the resulting wavelet coefficients separately and concatenating them into a matrix. This method allows us to utilize both time-domain and frequency-domain information simultaneously to capture the key features within the data, thereby enhancing the accuracy of predictions.
    \item We introduce a differential attention mechanism in time series forecasting, which computes the attention score by taking the difference between two separate $\mathrm{softmax}$ attention matrices. This can eliminate attention noise and enhance the focus on key information.
    \item Based on the above two designs, this paper proposes a novel time series forecasting framework, WDformer. The WDformer framework provides an efficient and accurate solution for time series forecasting by integrating the wavelet transform and the differential attention mechanism. In the experiments, WDformer achieves state-of-the-art (SOTA) results on multiple challenging real-world datasets.
\end{itemize}

\section{Related Work}

 The history of time series forecasting tasks is long and storied, initially relying heavily on traditional statistical methods such as Autoregressive models. However, with the rapid development of deep learning technology, this field has undergone profound changes. RNNs model leverages their recurrent structure to capture temporal dependencies in time series data for prediction~\cite{jia2024witran,luo2024moderntcn,wang2019deep}. TCNs/CNNs model effectively models data using the powerful feature extraction capabilities provided by convolution operations~\cite{wang2023micn,wu2022timesnet,sen2019think}. GNNs model, with their powerful graph structure representation capabilities through message passing mechanisms, effectively model time series data~\cite{cao2020spectral,cai2024msgnet,wu2019graph,wu2020connecting}. State Space Models, with their powerful ability to model dynamic systems through state transition mechanisms, can efficiently process time series data and make predictions~\cite{zhou2023deep,zhang2023effectively,li2021learning}. LLMs have also shown significant effectiveness in the field of time series forecasting~\cite{pan2024s,jin2023time,jin2024position}.

 Transformer models have garnered widespread attention due to their exceptional ability to capture long-distance dependencies. Informer~\cite{zhou2021informer} proposes the ProbSparse self-attention mechanism. Autoformer~\cite{wu2021autoformer} proposes an autocorrelation mechanism and sequence decomposition to effectively capture long-term dependencies in time series. FEDformer~\cite{zhou2022fedformer} combines the Fourier transform and Transformer, utilizing frequency domain sparsity to reduce computational complexity. PatchTST ~\cite{nie2022time}improves prediction accuracy through patching and channel independence. iTransformer~\cite{liu2023itransformer} enhances performance by applying attention and feed-forward networks to inverted dimensions.

 In time series forecasting, wavelet transform can assist in multi-scale feature extraction and noise suppression. \citet{jamal2021wavelet} proposed a hybrid model combining wavelet transform with neural networks, which optimizes the input of neural networks through wavelet decomposition. W-Transformers\cite{sasal2022w} handle long-range dependencies in non-stationary time series by integrating the Maximal Overlap Discrete Wavelet Transform (MODWT) with the Transformer architecture. WHEN \cite{wang2023wavelet} combines wavelet transform with Dynamic Time Warping (DTW) to improve the accuracy of time series forecasting. These studies demonstrate the important role of wavelet transform in enhancing the accuracy of time series forecasting.

\section{Model Structure}

Time series data typically exhibit temporal dependencies, where the value of a current data point is related to the values of past data points. Time series forecasting is based on predicting future data using past time series data. %Let's first understand 
Here is the formulation of the time series forecasting task: Given a time series sample $X = \left \{ {x}_{1},...,{x}_{K}\right \}\in {\mathbb{R}}^{K\times N}$ with a review window length of $\mathit{K}$ which means $\mathit{K}$ timestamps and $\mathit{N} $ variables. We need to predict the time series $Y = \left \{ {x}_{K+1},...,{x}_{K+F}\right \}\in {\mathbb{R}}^{F\times N}$ for the next $\mathit{F} $ timestamps. %moments.

\begin{figure*}[ht]
\centering
\includegraphics[width=1.0\textwidth,height=0.55\textheight]{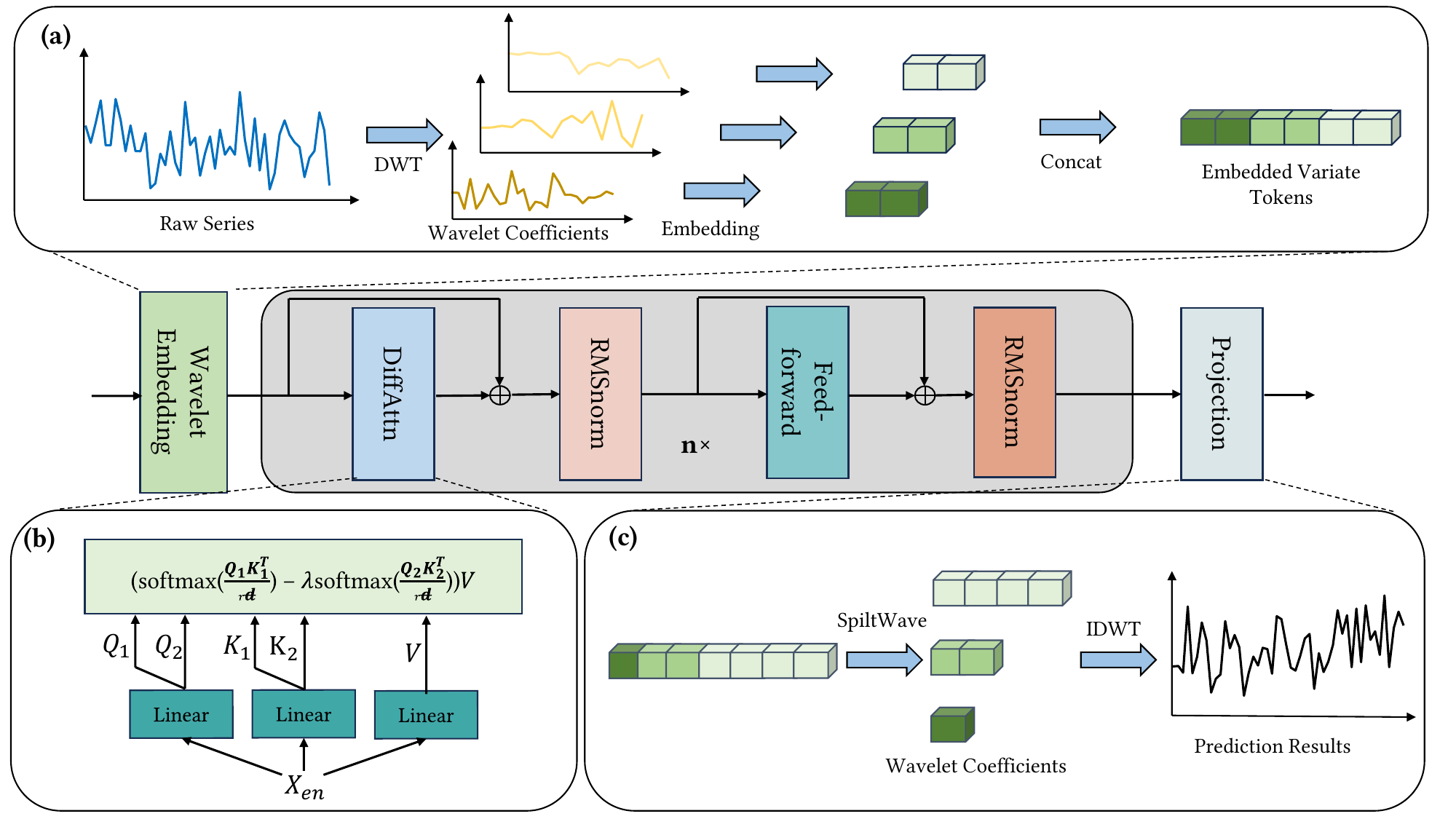}
\caption{The overall architecture of WDformer adopts an Encoder-only structure. (a) Wavelet transform is performed on the original time series data to extract features at different frequencies. Subsequently, each obtained wavelet coefficient undergoes independent embedding processing, and these embedding results are concatenated. (b) The differential attention mechanism uses the difference between two separate $\mathrm{softmax}$ attention matrices as the attention score. (c) The output results are reasonably segmented according to the length, which allows for the Inverse Discrete Wavelet Transform (IDWT). Subsequently, the IDWT technique is employed to reconstruct the predicted results of the time series data.}
\label{figure}
\end{figure*}

\subsection{Overall Model Architecture}

The WDformer structure we propose is based on an encoder-only transformer architecture, with modifications to the embedding, attention mechanism. The Transformer architecture was initially designed for NLP tasks, with its encoder-decoder structure tightly coupled with the autoregressive generation process. However, in the field of time series forecasting, a powerful encoder combined with a lightweight prediction head has proven to be a simple and effective solution\cite{liu2023itransformer, nie2022time}. This not only improves training efficiency but also avoids autoregressive decoding errors. Therefore, we have chosen to adopt an encoder-only architecture. The general structure is shown in Figure~\ref{figure}, which is primarily composed of wavelet embedding, differential attention, feed-forward, and projection. Most previous transformer-based models primarily used attention mechanisms to focus on temporal correlations. iTransformer~\cite{liu2023itransformer} treats independent time series as tokens and uses a self-attention mechanism to capture the multivariate correlations among them, but this approach only focuses on temporal information and neglects frequency domain information. The model we propose improves upon this by performing a Discrete Wavelet Transform(DWT) on the original time series, thereby enabling the attention mechanism to focus on both temporal and frequency-domain information at the same time. Subsequently, we employ differential attention to reduce noise. This mechanism is particularly important in time series forecasting because time series data often contains a large amount of noise, especially in long sequences and non-stationary data. The differential attention mechanism can effectively suppress the model's attention to irrelevant contexts, thereby improving the accuracy and robustness of predictions. The specific details are presented in the text below.

\subsection{Wavelet Embedding}
To simultaneously leverage information from both the time and frequency domains, WDformer first applies DWT to the raw data. Since different time series datasets have varying proportions of high and low-frequency information, using a fixed number of wavelet decomposition levels could either fail to effectively remove noise or inadvertently eliminate useful information. Therefore, we introduce a hyperparameter to adjust the number of wavelet transformation levels.

After applying $\mathit{L}$-level of DWT to the time series, a total of $\mathit{L}+1$ sets of wavelet coefficients are obtained.\@ To better capture the intrinsic structure and semantic information of the data, WDformer separately embeds each wavelet coefficient. The mathematical formula is expressed as follows:

\begin{align}
\mathrm{DWT}\left ( X \right ) =\left \{  \phi _{1},\phi _{2},...,\phi _{L+1}\right \} 
\end{align}%
\\
where $\mathit{X} $ represents the original time series data.\@ $\mathrm{DWT\left ( \cdot  \right ) }$ represents the DWT function. $\left \{  \phi _{1},\phi _{2},...,\phi _{L+1}\right \}$ represents the wavelet coefficients obtained after applying $\mathit{L} $-level of DWT to the original time series data. Next, we need to perform embedding operations independently on each wavelet coefficient and simply concatenate the results together.

\begin{align}
\hat{\phi _{i} }  = \mathrm{Embedding}\left (  \phi _{i}\right ) \quad i\in \left \{ 1,2,...,L+1 \right \}    
\end{align}%
\begin{align}
X_{\mathrm{en}}  = \mathrm{Concat} \left (   \hat{\phi}_{1} ,\hat{\phi}_{2},...,\hat{\phi }_{L+1}   \right )  
\end{align}%
\\
where $\mathrm{Embedding\left ( \cdot  \right ) }$ operation is to map the first $\mathit{L} $ wavelet coefficients to $\frac{d}{L+1} $, and $\hat{\phi }_{L+1}$ is mapped to $d - L\times \left ( \frac{d }{L+1} \right ) $, where d represents the dimension of the embedding vector. $X_{\mathrm{en}}\in {\mathbb{R}}^{N\times d}$ represents the output after Wavelet Embedding.

\subsection{Differential Attention}

The differential attention mechanism points out that when calculating the $\mathrm{softmax}$ attention matrix, the traditional attention calculation method often overly focuses on irrelevant information, thus weakening the model's ability to capture key features. To solve this problem, the differential attention mechanism separately calculates two independent $\mathrm{softmax}$ attention matrices, extracts more distinctive features by subtracting them, and finally obtains more accurate attention results. The following are the specific implementation details:

\begin{equation}
    \resizebox{.91\linewidth}{!}{$
            \displaystyle
            \left [ Q_{1};Q_{2}  \right ] = X_{\mathrm{en}}W^{Q} ,
\left [ K_{1};K_{2}  \right ] = X_{\mathrm{en}}W^{K} ,
V = X_{\mathrm{en}}W^{V}
        $}
\end{equation}%
\\
where $W^{Q},W^{K},W^{V}\in {\mathbb{R}}^{d\times 2d}$ are parameters, map the input $X_{\mathrm{en}}$  into query, key, and value,  respectively.\@ In the formula,
 $Q_1, Q_2, K_1,$ $  K_2 \in \mathbb{R}^{N \times d},\quad V \in \mathbb{R}^{N \times 2d}$ represent the query, key, and value, which will be used in our subsequent attention computation. The differential attention calculation formula $\mathrm{DiffAttn\left ( \cdot  \right ) }$ is defined as follows:

\begin{equation}
    \resizebox{.91\linewidth}{!}{$
            \displaystyle
            \mathrm{DiffAttn} \left ( X_{\mathrm{en}} \right ) = \left ( \mathrm{softmax} \left ( \frac{Q_{1}K_{1}^{T}  }{\sqrt{d} }  \right )-\lambda \mathrm{softmax} \left ( 
\frac{Q_{2}K_{2}^{T}  }{\sqrt{d} }  \right )  \right )V
        $}
\end{equation}%
\begin{align}
\lambda = \exp(\lambda _{q_{1} }  \cdot \lambda _{k_{1} })  - \exp(\lambda _{q_{2} }  \cdot \lambda _{k_{2} } ) + \lambda_{\mathrm{init}}
\end{align}%
\\
where $\lambda _{q_{1} } ,\lambda _{q_{2} } ,\lambda _{k_{1} } ,\lambda _{k_{2} } \in {\mathbb{R}}^{d}$ are learnable parameter and $\lambda _{\mathrm{init}  } $ is an initialized constant. Through experiments, we found that setting it to $0.7 - 0.5\times \exp \left ( -0.3\cdot \left ( l-1 \right )  \right ) $ is better in time series forecasting tasks. where $\mathit{l} $ is the current number of encoder layers.

\textbf{Multi-Head Differential Attention} The multi-head mechanism in the differential attention mechanism not only inherits many advantages of the traditional multi-head attention mechanism but also deeply integrates with the core concept of differential attention. The differential attention mechanism implements the differential attention calculation in each independent head to capture the key features in the input data and reduce the interference of irrelevant information. After the calculations of all the heads are completed, the output results of these heads will be spliced together along the channel dimension through the concat function to form a comprehensive feature representation. To further improve the training stability of the model and optimize the learning process, we perform normalization operations on the output of each head. The specific details are as follows:

\begin{align}
\overline{\mathrm{head} _{k} }  = \left ( 1-\lambda _{\mathrm{init} }  \right ) \cdot \mathrm{LN} \left (  \mathrm{head} _{k}\right )  
\end{align}%
\begin{equation}
    \resizebox{.91\linewidth}{!}{$
            \displaystyle
            \mathrm{MultiHead} \left ( X_{\mathrm{en}}  \right ) = \mathrm{Concat}\left ( \overline{\mathrm{head} _{1} },\overline{\mathrm{head} _{2} },...,\overline{\mathrm{head} _{h} } \right ) 
        $}
\end{equation}%
\\
where $\mathrm{head} _{k}$ represents the $k$-th head that has completed the attention calculation, $\overline{\mathrm{head} _{k} }$ represents the result after completing the normalization, and h represents the hyperparameter representing the number of heads we set. It is worth noting that the variable $\lambda _{\mathrm{init} }$ is the same variable when calculating the differential attention. To ensure the stability and efficiency of the model training, $\mathrm{LN \left ( \cdot  \right )}$ uses RMSNorm for each head. This normalization method helps adjust the output of each head to have a unified scale, thereby promoting more effective learning and convergence.

\subsection{Prediction} % 要用名词
After multiple layers of encoder training, we now obtain the output $X_{\mathrm{out}} $. WDformer adopts an encoder-only mode and performs a simple linear layer calculation on the output to obtain the predicted time series wavelet coefficients with $\hat{Y} $.

\begin{align}
\hat{Y} = \mathrm{FeedForward} \left ( X_{\mathrm{out}} \right ) 
\end{align}%
\\
where $\hat{Y}\in {\mathbb{R}}^{N\times F}$ , $F$ represents the length of the time series forecast. We decompose  $\hat{Y}$ to obtain the wavelet coefficients for the IDWT. The wavelet coefficients are mainly divided into approximation coefficients and detail coefficients. In the multi-level wavelet transform, the approximation coefficients and detail coefficients at each level are obtained through further decomposition of the approximation coefficients from the previous level. The lengths of these coefficients are not equal, so during the inverse wavelet transform, we cannot simply split the data equally. Instead, we need to gradually reconstruct the data according to the length and structure of the coefficients at each level and in the reverse process of the wavelet transform, to ensure that the recovered signal is accurate and correct. The output $\hat{Y}$ that we previously obtained is the concatenation of the wavelet coefficients from multiple layers. We need to split it. The definition of the split function $\mathrm{SpiltWave \left ( \cdot  \right )}$ is as follows:

\begin{align}
\mathrm{SpiltWave} \left ( \hat{Y} \right )=\left \{ \hat{Y_{1}},\hat{Y_{2}},\hat{Y_{3}},...,\hat{Y}_{L+1}  \right \} 
\end{align}%
\\
where $\hat{Y_{1}} $ represents the approximation coefficients and $\hat{Y_{j}},j \in \left ( 1,2,..., L+1 \right ) $ represents the coefficient of our predicted time series data. The $\hat{Y_{1}} $ and $\hat{Y_{2}}$ respectively represent the approximation coefficient and coefficient of the last level, and the length of their dimensions used for the inverse wavelet transform should both be $\frac{F}{2^{L} } $, where $F$ is the forecasting length and $L$ is the number of levels of the wavelet transform. When performing the inverse wavelet transform, what we first need to make clear is that the inverse process of the wavelet transform requires us to start from the approximation coefficients and coefficients of the lowest level and gradually reconstruct upwards until reaching the length of the original signal. Therefore, when we start the inverse transform from the lowest level, the length of the coefficients at each level should be twice that of the coefficients of the previous level. After the split is completed, we can perform the inverse wavelet transform on the split data.

\begin{align}
Y = \mathrm{IDWT} \left ( \mathrm{SpiltWave} \left ( \hat{Y} \right ) \right ) 
\end{align}%
\\
 $\mathrm{IDWT\left ( \cdot  \right ) }$ represents the IDWT. After performing the IDWT, we obtained the final prediction result $\mathit{Y} $.

\section{Experiments}

% aiming not aimed
In this section, we design and carry out a wide range of experiments aiming at comprehensively evaluating the predictive performance of the proposed model. %relative to the current state-of-the-art forecasting models.

\begin{table*}[htbp]
\resizebox{\linewidth}{!}{
\begin{tabular}{@{}l|llllllllllllllllll@{}}
\toprule
Models   & \multicolumn{2}{c}{WDformer(Ours)}                                           & \multicolumn{2}{c}{iTransformer}                                             & \multicolumn{2}{c}{PatchTST}                                                 & \multicolumn{2}{c}{Crossformer}    & \multicolumn{2}{c}{TiDE}           & \multicolumn{2}{c}{TimesNet}       & \multicolumn{2}{c}{Dlinear}                             & \multicolumn{2}{c}{FEDformer}      & \multicolumn{2}{c}{Autoformer} \\ \midrule
Metric   & MSE                        & \multicolumn{1}{l|}{MAE}                        & MSE                        & \multicolumn{1}{l|}{MAE}                        & MSE                        & \multicolumn{1}{l|}{MAE}                        & MSE   & \multicolumn{1}{l|}{MAE}   & MSE   & \multicolumn{1}{l|}{MAE}   & MSE   & \multicolumn{1}{l|}{MAE}   & MSE                        & \multicolumn{1}{l|}{MAE}   & MSE   & \multicolumn{1}{l|}{MAE}   & MSE            & MAE           \\ \midrule
ECL      & \color[HTML]{FE0000} 0.171 & \multicolumn{1}{l|}{\color[HTML]{FE0000} 0.263} & \color[HTML]{3531FF} 0.178 & \multicolumn{1}{l|}{\color[HTML]{3531FF} 0.270} & 0.205                      & \multicolumn{1}{l|}{0.290}                      & 0.244 & \multicolumn{1}{l|}{0.344} & 0.192 & \multicolumn{1}{l|}{0.295} & 0.192 & \multicolumn{1}{l|}{0.295} & 0.212                      & \multicolumn{1}{l|}{0.300} & 0.214 & \multicolumn{1}{l|}{0.327} & 0.227          & 0.338         \\ \midrule
ETT(avg) & 0.386                      & \multicolumn{1}{l|}{0.401}                      & \color[HTML]{3531FF} 0.383 & \multicolumn{1}{l|}{\color[HTML]{3531FF} 0.399} & \color[HTML]{FE0000} 0.381 & \multicolumn{1}{l|}{\color[HTML]{FE0000} 0.397} & 0.685 & \multicolumn{1}{l|}{0.578} & 0.482 & \multicolumn{1}{l|}{0.470} & 0.391 & \multicolumn{1}{l|}{0.404} & 0.442                      & \multicolumn{1}{l|}{0.444} & 0.408 & \multicolumn{1}{l|}{0.428} & 0.465          & 0.459         \\ \midrule
Exchange & \color[HTML]{3531FF} 0.358 & \multicolumn{1}{l|}{0.407}                      & 0.360                      & \multicolumn{1}{l|}{\color[HTML]{FE0000} 0.403} & 0.367                      & \multicolumn{1}{l|}{\color[HTML]{3531FF} 0.404} & 0.940 & \multicolumn{1}{l|}{0.707} & 0.370 & \multicolumn{1}{l|}{0.413} & 0.416 & \multicolumn{1}{l|}{0.443} & \color[HTML]{FE0000} 0.354 & \multicolumn{1}{l|}{0.414} & 0.519 & \multicolumn{1}{l|}{0.429} & 0.613          & 0.539         \\ \midrule
Traffic  & \color[HTML]{FE0000} 0.418 & \multicolumn{1}{l|}{\color[HTML]{FE0000} 0.279} & \color[HTML]{3531FF} 0.428 & \multicolumn{1}{l|}{\color[HTML]{3531FF} 0.282} & 0.481                      & \multicolumn{1}{l|}{0.304}                      & 0.550 & \multicolumn{1}{l|}{0.304} & 0.760 & \multicolumn{1}{l|}{0.473} & 0.620 & \multicolumn{1}{l|}{0.336} & 0.625                      & \multicolumn{1}{l|}{0.383} & 0.610 & \multicolumn{1}{l|}{0.376} & 0.628          & 0.379         \\ \midrule
Weather  & \color[HTML]{FE0000} 0.258 & \multicolumn{1}{l|}{\color[HTML]{3531FF} 0.279} & \color[HTML]{3531FF} 0.258 & \multicolumn{1}{l|}{\color[HTML]{FE0000} 0.278} & 0.259                      & \multicolumn{1}{l|}{0.281}                      & 0.259 & \multicolumn{1}{l|}{0.315} & 0.271 & \multicolumn{1}{l|}{0.320} & 0.259 & \multicolumn{1}{l|}{0.287} & 0.265                      & \multicolumn{1}{l|}{0.317} & 0.309 & \multicolumn{1}{l|}{0.360} & 0.338          & 0.382         \\ \bottomrule
\end{tabular}
}
\caption{Experiments on multivariate long-term sequence forecasting are conducted on five datasets with an input length of $K=96$ and prediction lengths of $F\in \left \{ 96, 192, 336, 720 \right \} $, and the average values across all prediction lengths were calculated. Lower MSE and MAE values indicate better predictive performance of the model. The optimal results are indicated in \textcolor[HTML]{FF0000}{red}, while the suboptimal results are shown in \textcolor[HTML]{3166FF}{blue}. In the experiment, a random factor was set, and the results were consistent each time it was run. There was no need to run it multiple times to take an average, and the same was true for other experiments.}
\label{tab:results_avg}
\end{table*}

% \usepackage{multirow}

\subsection{Experimental Settings}

\subsubsection{Datasets}

We conduct detailed tests on our model across eight multivariate time series datasets from the real world, including electricity, transportation, weather, and finance. They include ECL, ETT (ETTh1, ETTh2, ETTm1, ETTm2), Weather, Traffic, and Exchange.

\subsubsection{Evaluation Metrics}

Evaluation metrics are used to quantify the accuracy of model predictions and are key in measuring model performance. In time series forecasting tasks, commonly used evaluation metrics include Mean Squared Error (MSE) and Mean Absolute Error (MAE). Therefore, we also use these two metrics to evaluate our model. When evaluating the performance of a forecasting model, the smaller the values of Mean Absolute Error (MAE) and Mean Squared Error (MSE), the higher the accuracy of the forecasting results.

\subsubsection{Baselines}
We choose eight models as baselines, which are iTransformer~\cite{liu2023itransformer}, PatchTST~\cite{nie2022time}, Crossformer\cite{zhang2023crossformer}, TiDE\cite{das2023long}, TimesNet~\cite{wu2022timesnet}, Dlinear~\cite{zeng2023transformers}, FEDformer~\cite{zhou2022fedformer}, and Autoformer~\cite{wu2021autoformer}. To ensure fairness in experimental results, we strictly control variables. The  input length $K$ is set to 96, and the prediction length 
$F$ is set to {96, 192, 336, 720} to demonstrate the accuracy at different prediction lengths.

\subsection{Main Results}

As shown in Table \ref{tab:results_avg}, the mean value of the comprehensive forecasting results is indicated. The WDformer model demonstrates outstanding performance on the ECL and Traffic data sets, achieving the lowest Mean Absolute Error (MAE) and Mean Squared Error (MSE). Particularly on the ECL dataset, compared with the second-best model iTransformer, WDformer achieves a 3.9\% reduction in MSE and a 2.6\% reduction in MAE. On the other data sets, the WDformer also performs excellently, obtaining 1 optimal result and 2 sub-optimal results. 

It is worth noting that the ECL and Traffic data sets have significant periodic characteristics, and the amplitude of their peaks is relatively small. 
\begin{figure}[htbp]
    \centering
    % 第一行子图
    \begin{subfigure}[b]{0.23\textwidth}
        \centering
    \includegraphics[height=3.0cm,width=1\textwidth]{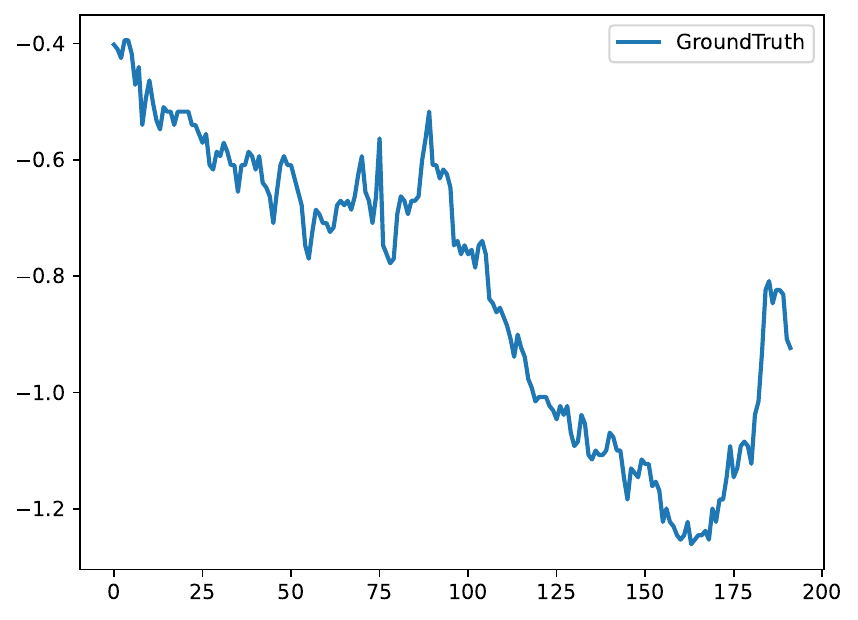}
        \caption{ETTm1}
        \label{ETTm1}
    \end{subfigure}
    \hfill % 添加间距
    \begin{subfigure}[b]{0.23\textwidth}
        \centering
    \includegraphics[height=3.0cm,width=1\textwidth]{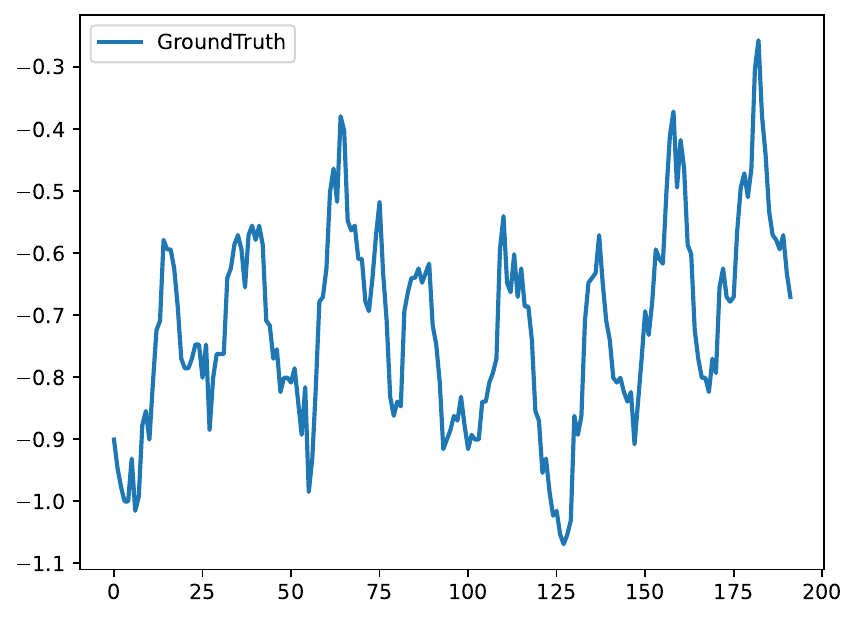}
    \caption{ETTh1}
        \label{ETTh1}
    \end{subfigure}
    \caption{Visualization of Time Series in the ETTh1 and ETTh2 Datasets}
    \label{figure:ETT}
\end{figure}
The main reason why the WDformer model can achieve such a significant advantage on these two data sets lies in the multi-scale feature extraction capability of its wavelet transform. This capability enables the model to efficiently and accurately identify and capture the hidden periodic patterns in the data, thereby providing a solid foundation for subsequent predictions. The differential attention mechanism adopted in the WDformer model also plays an important role. This mechanism further strengthens the model's focus on key information, enabling it to pay more attention to important information when processing data, and thus effectively improving the accuracy of predictions. 
On the ETT data set, the PatchTST model~\cite{nie2022time} achieves the optimal results. This advantage is mainly attributed to its unique channel-independent strategy. This strategy enables the model to more flexibly adapt to the characteristics of data sets like ETT, which are periodic, complex, and have large amplitude variations in peaks, giving full play to its advantages in complex scenarios. However, this strategy also leads to the problem of insufficient capture of the correlation between features when PatchTST faces relatively stable data sets like ECL and Traffic, thereby limiting its performance on these data sets. 

\begin{table*}[htbp]
\resizebox{\linewidth}{!}{
\begin{tabular}{@{}llllllllllllllllllll@{}}
\toprule
\multicolumn{2}{l|}{Models}                                                 & \multicolumn{2}{l}{WDformer(Ours)}                                               & \multicolumn{2}{l}{iTransformer}                                                 & \multicolumn{2}{c}{PatchTST}                                                     & \multicolumn{2}{c}{Crossformer}                             & \multicolumn{2}{c}{TiDE}                             & \multicolumn{2}{c}{TimesNet}                                                     & \multicolumn{2}{c}{Dlinear}                                                      & \multicolumn{2}{c}{FEDformer}                             & \multicolumn{2}{l}{Autoformer} \\ \midrule
\multicolumn{2}{l|}{Metric}                                                & MSE                          & \multicolumn{1}{l|}{MAE}                          & MSE                          & \multicolumn{1}{l|}{MAE}                          & MSE                          & \multicolumn{1}{l|}{MAE}                          & MSE                          & \multicolumn{1}{l|}{MAE}   & MSE                          & \multicolumn{1}{l|}{MAE}   & MSE                          & \multicolumn{1}{l|}{MAE}                          & MSE                          & \multicolumn{1}{l|}{MAE}                          & MSE                          & \multicolumn{1}{l|}{MAE}   & MSE            & MAE           \\ \midrule
\multicolumn{1}{l|}{}                           & \multicolumn{1}{l|}{96}  & 0.339                        & \multicolumn{1}{l|}{0.373}                        & {\color[HTML]{3531FF} 0.334} & \multicolumn{1}{l|}{{\color[HTML]{3531FF} 0.368}} & {\color[HTML]{FE0000} 0.329} & \multicolumn{1}{l|}{{\color[HTML]{FF0000} 0.367}} & 0.404                        & \multicolumn{1}{l|}{0.426} & 0.364                        & \multicolumn{1}{l|}{0.387} & 0.338                        & \multicolumn{1}{l|}{0.375}                        & 0.345                        & \multicolumn{1}{l|}{0.372}                        & 0.379                        & \multicolumn{1}{l|}{0.419} & 0.505          & 0.475         \\
\multicolumn{1}{l|}{}                           & \multicolumn{1}{l|}{192} & 0.384                        & \multicolumn{1}{l|}{0.399}                        & 0.377                        & \multicolumn{1}{l|}{0.391}                        & {\color[HTML]{FE0000} 0.367} & \multicolumn{1}{l|}{{\color[HTML]{FF0000} 0.385}} & 0.450                        & \multicolumn{1}{l|}{0.451} & 0.398                        & \multicolumn{1}{l|}{0.404} & {\color[HTML]{3531FF} 0.374} & \multicolumn{1}{l|}{{\color[HTML]{3531FF} 0.387}} & 0.380                        & \multicolumn{1}{l|}{0.389}                        & 0.426                        & \multicolumn{1}{l|}{0.441} & 0.553          & 0.496         \\
\multicolumn{1}{l|}{}                           & \multicolumn{1}{l|}{336} & 0.422                        & \multicolumn{1}{l|}{0.423}                        & 0.426                        & \multicolumn{1}{l|}{0.420}                        & {\color[HTML]{FE0000} 0.399} & \multicolumn{1}{l|}{{\color[HTML]{FF0000} 0.410}} & 0.532                        & \multicolumn{1}{l|}{0.515} & 0.428                        & \multicolumn{1}{l|}{0.425} & {\color[HTML]{3531FF} 0.410} & \multicolumn{1}{l|}{{\color[HTML]{3531FF} 0.411}} & 0.413                        & \multicolumn{1}{l|}{0.413}                        & 0.445                        & \multicolumn{1}{l|}{0.459} & 0.621          & 0.537         \\
\multicolumn{1}{l|}{\multirow{-4}{*}{ETTm1}}    & \multicolumn{1}{l|}{720} & 0.489                        & \multicolumn{1}{l|}{0.461}                        & 0.491                        & \multicolumn{1}{l|}{0.459}                        & {\color[HTML]{FE0000} 0.454} & \multicolumn{1}{l|}{{\color[HTML]{FF0000} 0.439}} & 0.666                        & \multicolumn{1}{l|}{0.589} & 0.487                        & \multicolumn{1}{l|}{0.461} & 0.478                        & \multicolumn{1}{l|}{{\color[HTML]{3531FF} 0.450}} & {\color[HTML]{3531FF} 0.474} & \multicolumn{1}{l|}{0.453}                        & 0.543                        & \multicolumn{1}{l|}{0.490} & 0.671          & 0.561         \\ \midrule
\multicolumn{1}{l|}{}                           & \multicolumn{1}{l|}{96}  & 0.181                        & \multicolumn{1}{l|}{{\color[HTML]{3531FF} 0.263}} & {\color[HTML]{3531FF} 0.180} & \multicolumn{1}{l|}{0.264}                        & {\color[HTML]{FE0000} 0.175} & \multicolumn{1}{l|}{{\color[HTML]{FF0000} 0.259}} & 0.287                        & \multicolumn{1}{l|}{0.366} & 0.207                        & \multicolumn{1}{l|}{0.305} & 0.187                        & \multicolumn{1}{l|}{0.267}                        & 0.193                        & \multicolumn{1}{l|}{0.292}                        & 0.203                        & \multicolumn{1}{l|}{0.287} & 0.255          & 0.339         \\
\multicolumn{1}{l|}{}                           & \multicolumn{1}{l|}{192} & {\color[HTML]{3531FF} 0.247} & \multicolumn{1}{l|}{{\color[HTML]{3531FF} 0.305}} & 0.250                        & \multicolumn{1}{l|}{0.309}                        & {\color[HTML]{FE0000} 0.241} & \multicolumn{1}{l|}{{\color[HTML]{FF0000} 0.302}} & 0.414                        & \multicolumn{1}{l|}{0.492} & 0.290                        & \multicolumn{1}{l|}{0.364} & 0.249                        & \multicolumn{1}{l|}{0.309}                        & 0.284                        & \multicolumn{1}{l|}{0.362}                        & 0.269                        & \multicolumn{1}{l|}{0.328} & 0.281          & 0.340         \\
\multicolumn{1}{l|}{}                           & \multicolumn{1}{l|}{336} & 0.314                        & \multicolumn{1}{l|}{0.350}                        & {\color[HTML]{3531FF} 0.311} & \multicolumn{1}{l|}{{\color[HTML]{3531FF} 0.348}} & {\color[HTML]{FE0000} 0.305} & \multicolumn{1}{l|}{{\color[HTML]{FF0000} 0.343}} & 0.597                        & \multicolumn{1}{l|}{0.542} & 0.377                        & \multicolumn{1}{l|}{0.422} & 0.321                        & \multicolumn{1}{l|}{0.351}                        & 0.369                        & \multicolumn{1}{l|}{0.427}                        & 0.325                        & \multicolumn{1}{l|}{0.366} & 0.339          & 0.372         \\
\multicolumn{1}{l|}{\multirow{-4}{*}{ETTm2}}    & \multicolumn{1}{l|}{720} & 0.417                        & \multicolumn{1}{l|}{0.408}                        & 0.412                        & \multicolumn{1}{l|}{0.407}                        & {\color[HTML]{FE0000} 0.402} & \multicolumn{1}{l|}{{\color[HTML]{FF0000} 0.400}} & 1.730                        & \multicolumn{1}{l|}{1.042} & 0.558                        & \multicolumn{1}{l|}{0.524} & {\color[HTML]{3531FF} 0.408} & \multicolumn{1}{l|}{{\color[HTML]{3531FF} 0.403}} & 0.554                        & \multicolumn{1}{l|}{0.522}                        & 0.421                        & \multicolumn{1}{l|}{0.415} & 0.433          & 0.432         \\ \midrule
\multicolumn{1}{l|}{}                           & \multicolumn{1}{l|}{96}  & 0.391                        & \multicolumn{1}{l|}{0.409}                        & 0.386                        & \multicolumn{1}{l|}{0.405}                        & 0.414                        & \multicolumn{1}{l|}{0.419}                        & 0.423                        & \multicolumn{1}{l|}{0.448} & 0.479                        & \multicolumn{1}{l|}{0.464} & {\color[HTML]{3531FF} 0.384} & \multicolumn{1}{l|}{{\color[HTML]{3531FF} 0.402}} & 0.386                        & \multicolumn{1}{l|}{{\color[HTML]{FF0000} 0.400}} & {\color[HTML]{FE0000} 0.376} & \multicolumn{1}{l|}{0.419} & 0.449          & 0.459         \\
\multicolumn{1}{l|}{}                           & \multicolumn{1}{l|}{192} & 0.445                        & \multicolumn{1}{l|}{0.438}                        & 0.441                        & \multicolumn{1}{l|}{0.436}                        & 0.460                        & \multicolumn{1}{l|}{0.445}                        & 0.471                        & \multicolumn{1}{l|}{0.474} & 0.525                        & \multicolumn{1}{l|}{0.492} & {\color[HTML]{3531FF} 0.436} & \multicolumn{1}{l|}{{\color[HTML]{FF0000} 0.429}} & 0.437                        & \multicolumn{1}{l|}{{\color[HTML]{3531FF} 0.432}} & {\color[HTML]{FE0000} 0.420} & \multicolumn{1}{l|}{0.448} & 0.500          & 0.482         \\
\multicolumn{1}{l|}{}                           & \multicolumn{1}{l|}{336} & 0.491                        & \multicolumn{1}{l|}{0.466}                        & 0.487                        & \multicolumn{1}{l|}{{\color[HTML]{FF0000} 0.458}} & 0.501                        & \multicolumn{1}{l|}{0.466}                        & 0.570                        & \multicolumn{1}{l|}{0.546} & 0.565                        & \multicolumn{1}{l|}{0.515} & 0.491                        & \multicolumn{1}{l|}{0.469}                        & {\color[HTML]{3531FF} 0.481} & \multicolumn{1}{l|}{{\color[HTML]{3531FF} 0.459}} & {\color[HTML]{FE0000} 0.459} & \multicolumn{1}{l|}{0.465} & 0.521          & 0.496         \\
\multicolumn{1}{l|}{\multirow{-4}{*}{ETTh1}}    & \multicolumn{1}{l|}{720} & 0.514                        & \multicolumn{1}{l|}{0.496}                        & {\color[HTML]{3531FF} 0.503} & \multicolumn{1}{l|}{{\color[HTML]{3531FF} 0.491}} & {\color[HTML]{FF0000} 0.500} & \multicolumn{1}{l|}{{\color[HTML]{FF0000} 0.488}} & 0.653                        & \multicolumn{1}{l|}{0.621} & 0.594                        & \multicolumn{1}{l|}{0.558} & 0.521                        & \multicolumn{1}{l|}{0.500}                        & 0.519                        & \multicolumn{1}{l|}{0.516}                        & 0.506                        & \multicolumn{1}{l|}{0.507} & 0.514          & 0.512         \\ \midrule
\multicolumn{1}{l|}{}                           & \multicolumn{1}{l|}{96}  & {\color[HTML]{3531FF} 0.302} & \multicolumn{1}{l|}{{\color[HTML]{3531FF} 0.349}} & {\color[HTML]{FF0000} 0.297} & \multicolumn{1}{l|}{0.349}                        & 0.302                        & \multicolumn{1}{l|}{{\color[HTML]{FF0000} 0.348}} & 0.745                        & \multicolumn{1}{l|}{0.584} & 0.400                        & \multicolumn{1}{l|}{0.440} & 0.340                        & \multicolumn{1}{l|}{0.374}                        & 0.333                        & \multicolumn{1}{l|}{0.387}                        & 0.358                        & \multicolumn{1}{l|}{0.397} & 0.346          & 0.388         \\
\multicolumn{1}{l|}{}                           & \multicolumn{1}{l|}{192} & {\color[HTML]{3531FF} 0.382} & \multicolumn{1}{l|}{0.402}                        & {\color[HTML]{FF0000} 0.380} & \multicolumn{1}{l|}{{\color[HTML]{FF0000} 0.400}} & 0.388                        & \multicolumn{1}{l|}{{\color[HTML]{3531FF} 0.400}} & 0.877                        & \multicolumn{1}{l|}{0.656} & 0.528                        & \multicolumn{1}{l|}{0.509} & 0.402                        & \multicolumn{1}{l|}{0.414}                        & 0.477                        & \multicolumn{1}{l|}{0.476}                        & 0.429                        & \multicolumn{1}{l|}{0.439} & 0.456          & 0.452         \\
\multicolumn{1}{l|}{}                           & \multicolumn{1}{l|}{336} & {\color[HTML]{FF0000} 0.425} & \multicolumn{1}{l|}{{\color[HTML]{FF0000} 0.432}} & 0.428                        & \multicolumn{1}{l|}{{\color[HTML]{3531FF} 0.432}} & {\color[HTML]{3531FF} 0.426} & \multicolumn{1}{l|}{0.433}                        & 1.043                        & \multicolumn{1}{l|}{0.731} & 0.643                        & \multicolumn{1}{l|}{0.571} & 0.452                        & \multicolumn{1}{l|}{0.452}                        & 0.594                        & \multicolumn{1}{l|}{0.541}                        & 0.496                        & \multicolumn{1}{l|}{0.487} & 0.482          & 0.486         \\
\multicolumn{1}{l|}{\multirow{-4}{*}{ETTh2}}    & \multicolumn{1}{l|}{720} & 0.433                        & \multicolumn{1}{l|}{{\color[HTML]{3531FF} 0.446}} & {\color[HTML]{FE0000} 0.427} & \multicolumn{1}{l|}{{\color[HTML]{FF0000} 0.445}} & {\color[HTML]{3531FF} 0.431} & \multicolumn{1}{l|}{0.446}                        & 1.104                        & \multicolumn{1}{l|}{0.763} & 0.871                        & \multicolumn{1}{l|}{0.679} & 0.462                        & \multicolumn{1}{l|}{0.468}                        & 0.831                        & \multicolumn{1}{l|}{0.657}                        & 0.463                        & \multicolumn{1}{l|}{0.474} & 0.515          & 0.511         \\ \midrule
\multicolumn{1}{l|}{}                           & \multicolumn{1}{l|}{96}  & {\color[HTML]{FE0000} 0.144} & \multicolumn{1}{l|}{{\color[HTML]{FE0000} 0.237}} & {\color[HTML]{3531FF} 0.148} & \multicolumn{1}{l|}{{\color[HTML]{3531FF} 0.240}} & 0.181                        & \multicolumn{1}{l|}{0.270}                        & 0.219                        & \multicolumn{1}{l|}{0.314} & 0.237                        & \multicolumn{1}{l|}{0.329} & 0.168                        & \multicolumn{1}{l|}{0.272}                        & 0.197                        & \multicolumn{1}{l|}{0.282}                        & 0.193                        & \multicolumn{1}{l|}{0.308} & 0.201          & 0.317         \\
\multicolumn{1}{l|}{}                           & \multicolumn{1}{l|}{192} & {\color[HTML]{FE0000} 0.161} & \multicolumn{1}{l|}{{\color[HTML]{FE0000} 0.253}} & {\color[HTML]{3531FF} 0.162} & \multicolumn{1}{l|}{{\color[HTML]{3531FF} 0.253}} & 0.188                        & \multicolumn{1}{l|}{0.274}                        & 0.231                        & \multicolumn{1}{l|}{0.322} & 0.236                        & \multicolumn{1}{l|}{0.330} & 0.184                        & \multicolumn{1}{l|}{0.289}                        & 0.196                        & \multicolumn{1}{l|}{0.285}                        & 0.201                        & \multicolumn{1}{l|}{0.315} & 0.222          & 0.334         \\
\multicolumn{1}{l|}{}                           & \multicolumn{1}{l|}{336} & {\color[HTML]{FE0000} 0.173} & \multicolumn{1}{l|}{{\color[HTML]{FE0000} 0.267}} & {\color[HTML]{3531FF} 0.178} & \multicolumn{1}{l|}{{\color[HTML]{3531FF} 0.269}} & 0.204                        & \multicolumn{1}{l|}{0.293}                        & 0.246                        & \multicolumn{1}{l|}{0.337} & 0.249                        & \multicolumn{1}{l|}{0.344} & 0.198                        & \multicolumn{1}{l|}{0.300}                        & 0.209                        & \multicolumn{1}{l|}{0.301}                        & 0.214                        & \multicolumn{1}{l|}{0.329} & 0.231          & 0.338         \\
\multicolumn{1}{l|}{\multirow{-4}{*}{ECL}}      & \multicolumn{1}{l|}{720} & {\color[HTML]{FE0000} 0.205} & \multicolumn{1}{l|}{{\color[HTML]{FE0000} 0.296}} & 0.225                        & \multicolumn{1}{l|}{{\color[HTML]{3531FF} 0.317}} & 0.246                        & \multicolumn{1}{l|}{0.324}                        & 0.280                        & \multicolumn{1}{l|}{0.363} & 0.284                        & \multicolumn{1}{l|}{0.373} & {\color[HTML]{3531FF} 0.220} & \multicolumn{1}{l|}{0.320}                        & 0.245                        & \multicolumn{1}{l|}{0.333}                        & 0.246                        & \multicolumn{1}{l|}{0.355} & 0.254          & 0.361         \\ \midrule
\multicolumn{1}{l|}{}                           & \multicolumn{1}{l|}{96}  & 0.089                        & \multicolumn{1}{l|}{0.210}                        & {\color[HTML]{FE0000} 0.086} & \multicolumn{1}{l|}{{\color[HTML]{3531FF} 0.206}} & 0.088                        & \multicolumn{1}{l|}{{\color[HTML]{FF0000} 0.205}} & 0.256                        & \multicolumn{1}{l|}{0.367} & 0.094                        & \multicolumn{1}{l|}{0.218} & 0.107                        & \multicolumn{1}{l|}{0.234}                        & {\color[HTML]{3531FF} 0.088} & \multicolumn{1}{l|}{0.218}                        & 0.148                        & \multicolumn{1}{l|}{0.278} & 0.197          & 0.323         \\
\multicolumn{1}{l|}{}                           & \multicolumn{1}{l|}{192} & 0.188                        & \multicolumn{1}{l|}{0.310}                        & 0.177                        & \multicolumn{1}{l|}{{\color[HTML]{3531FF} 0.299}} & {\color[HTML]{3531FF} 0.176} & \multicolumn{1}{l|}{{\color[HTML]{FF0000} 0.299}} & 0.470                        & \multicolumn{1}{l|}{0.509} & 0.184                        & \multicolumn{1}{l|}{0.307} & 0.226                        & \multicolumn{1}{l|}{0.344}                        & {\color[HTML]{FF0000} 0.176} & \multicolumn{1}{l|}{0.315}                        & 0.271                        & \multicolumn{1}{l|}{0.315} & 0.300          & 0.369         \\
\multicolumn{1}{l|}{}                           & \multicolumn{1}{l|}{336} & 0.337                        & \multicolumn{1}{l|}{0.422}                        & 0.331                        & \multicolumn{1}{l|}{{\color[HTML]{3531FF} 0.417}} & {\color[HTML]{FE0000} 0.301} & \multicolumn{1}{l|}{{\color[HTML]{FF0000} 0.397}} & 1.268                        & \multicolumn{1}{l|}{0.883} & 0.349                        & \multicolumn{1}{l|}{0.431} & 0.367                        & \multicolumn{1}{l|}{0.448}                        & {\color[HTML]{3531FF} 0.313} & \multicolumn{1}{l|}{0.427}                        & 0.460                        & \multicolumn{1}{l|}{0.427} & 0.509          & 0.524         \\
\multicolumn{1}{l|}{\multirow{-4}{*}{Exchange}} & \multicolumn{1}{l|}{720} & {\color[HTML]{FF0000} 0.817} & \multicolumn{1}{l|}{{\color[HTML]{FF0000} 0.685}} & 0.847                        & \multicolumn{1}{l|}{{\color[HTML]{3531FF} 0.691}} & 0.901                        & \multicolumn{1}{l|}{0.714}                        & 1.767                        & \multicolumn{1}{l|}{1.068} & 0.852                        & \multicolumn{1}{l|}{0.698} & 0.964                        & \multicolumn{1}{l|}{0.746}                        & {\color[HTML]{3531FF} 0.839} & \multicolumn{1}{l|}{0.695}                        & 1.195                        & \multicolumn{1}{l|}{0.695} & 1.447          & 0.941         \\ \midrule
\multicolumn{1}{l|}{}                           & \multicolumn{1}{l|}{96}  & {\color[HTML]{FE0000} 0.391} & \multicolumn{1}{l|}{{\color[HTML]{FE0000} 0.265}} & {\color[HTML]{3531FF} 0.395} & \multicolumn{1}{l|}{{\color[HTML]{3531FF} 0.268}} & 0.462                        & \multicolumn{1}{l|}{0.295}                        & 0.522                        & \multicolumn{1}{l|}{0.290} & 0.805                        & \multicolumn{1}{l|}{0.493} & 0.593                        & \multicolumn{1}{l|}{0.321}                        & 0.650                        & \multicolumn{1}{l|}{0.396}                        & 0.587                        & \multicolumn{1}{l|}{0.366} & 0.613          & 0.388         \\
\multicolumn{1}{l|}{}                           & \multicolumn{1}{l|}{192} & {\color[HTML]{FE0000} 0.403} & \multicolumn{1}{l|}{{\color[HTML]{FE0000} 0.271}} & {\color[HTML]{3531FF} 0.417} & \multicolumn{1}{l|}{{\color[HTML]{3531FF} 0.276}} & 0.466                        & \multicolumn{1}{l|}{0.296}                        & 0.530                        & \multicolumn{1}{l|}{0.293} & 0.756                        & \multicolumn{1}{l|}{0.474} & 0.617                        & \multicolumn{1}{l|}{0.336}                        & 0.598                        & \multicolumn{1}{l|}{0.370}                        & 0.604                        & \multicolumn{1}{l|}{0.373} & 0.616          & 0.382         \\
\multicolumn{1}{l|}{}                           & \multicolumn{1}{l|}{336} & {\color[HTML]{FE0000} 0.423} & \multicolumn{1}{l|}{{\color[HTML]{FE0000} 0.280}} & {\color[HTML]{3531FF} 0.433} & \multicolumn{1}{l|}{{\color[HTML]{3531FF} 0.283}} & 0.482                        & \multicolumn{1}{l|}{0.304}                        & 0.558                        & \multicolumn{1}{l|}{0.305} & 0.762                        & \multicolumn{1}{l|}{0.477} & 0.629                        & \multicolumn{1}{l|}{0.336}                        & 0.605                        & \multicolumn{1}{l|}{0.373}                        & 0.621                        & \multicolumn{1}{l|}{0.383} & 0.622          & 0.337         \\
\multicolumn{1}{l|}{\multirow{-4}{*}{Traffic}}  & \multicolumn{1}{l|}{720} & {\color[HTML]{FE0000} 0.456} & \multicolumn{1}{l|}{{\color[HTML]{FE0000} 0.298}} & {\color[HTML]{3531FF} 0.467} & \multicolumn{1}{l|}{{\color[HTML]{3531FF} 0.302}} & 0.514                        & \multicolumn{1}{l|}{0.322}                        & 0.589                        & \multicolumn{1}{l|}{0.328} & 0.719                        & \multicolumn{1}{l|}{0.449} & 0.640                        & \multicolumn{1}{l|}{0.350}                        & 0.645                        & \multicolumn{1}{l|}{0.394}                        & 0.626                        & \multicolumn{1}{l|}{0.382} & 0.660          & 0.408         \\ \midrule
\multicolumn{1}{l|}{}                           & \multicolumn{1}{l|}{96}  & {\color[HTML]{3531FF} 0.172} & \multicolumn{1}{l|}{{\color[HTML]{FF0000} 0.212}} & 0.174                        & \multicolumn{1}{l|}{{\color[HTML]{3531FF} 0.214}} & 0.177                        & \multicolumn{1}{l|}{0.218}                        & {\color[HTML]{FE0000} 0.158} & \multicolumn{1}{l|}{0.230} & 0.202                        & \multicolumn{1}{l|}{0.261} & 0.172                        & \multicolumn{1}{l|}{0.220}                        & 0.196                        & \multicolumn{1}{l|}{0.255}                        & 0.217                        & \multicolumn{1}{l|}{0.296} & 0.266          & 0.336         \\
\multicolumn{1}{l|}{}                           & \multicolumn{1}{l|}{192} & 0.224                        & \multicolumn{1}{l|}{{\color[HTML]{3531FF} 0.256}} & 0.221                        & \multicolumn{1}{l|}{{\color[HTML]{FE0000} 0.254}} & 0.225                        & \multicolumn{1}{l|}{0.259}                        & {\color[HTML]{FE0000} 0.206} & \multicolumn{1}{l|}{0.277} & 0.242                        & \multicolumn{1}{l|}{0.298} & {\color[HTML]{3531FF} 0.219} & \multicolumn{1}{l|}{0.261}                        & 0.237                        & \multicolumn{1}{l|}{0.296}                        & 0.276                        & \multicolumn{1}{l|}{0.336} & 0.307          & 0.367         \\
\multicolumn{1}{l|}{}                           & \multicolumn{1}{l|}{336} & 0.281                        & \multicolumn{1}{l|}{0.298}                        & {\color[HTML]{3531FF} 0.278} & \multicolumn{1}{l|}{{\color[HTML]{FE0000} 0.296}} & 0.278                        & \multicolumn{1}{l|}{{\color[HTML]{3531FF} 0.297}} & {\color[HTML]{FE0000} 0.272} & \multicolumn{1}{l|}{0.335} & 0.287                        & \multicolumn{1}{l|}{0.335} & 0.280                        & \multicolumn{1}{l|}{0.306}                        & 0.283                        & \multicolumn{1}{l|}{0.335}                        & 0.339                        & \multicolumn{1}{l|}{0.380} & 0.359          & 0.395         \\
\multicolumn{1}{l|}{\multirow{-4}{*}{Weather}}  & \multicolumn{1}{l|}{720} & 0.356                        & \multicolumn{1}{l|}{{\color[HTML]{3531FF} 0.348}} & 0.358                        & \multicolumn{1}{l|}{{\color[HTML]{FE0000} 0.347}} & 0.354                        & \multicolumn{1}{l|}{0.348}                        & 0.398                        & \multicolumn{1}{l|}{0.418} & {\color[HTML]{3531FF} 0.351} & \multicolumn{1}{l|}{0.386} & 0.365                        & \multicolumn{1}{l|}{0.359}                        & {\color[HTML]{FE0000} 0.345} & \multicolumn{1}{l|}{0.381}                        & 0.403                        & \multicolumn{1}{l|}{0.428} & 0.419          & 0.428         \\ \bottomrule
\end{tabular}}
\caption{The results of multivariate time series forecasting. Experiments on multivariate long-term sequence forecasting are conducted on eight datasets with an input length of $K=96$ and prediction lengths of $F\in \left \{ 96, 192, 336, 720 \right \} $. Lower MSE and MAE values indicate better predictive performance of the model. The optimal results are indicated in \textcolor[HTML]{FF0000}{red}, while the suboptimal results are shown in \textcolor[HTML]{3166FF}{blue}.}
\label{tab:results}
\end{table*}

In Table \ref{tab:results}, we show in detail the prediction results of each data set under different lengths. WDformer demonstrates exceptional performance on the ECL and Traffic datasets, achieving optimal results across all prediction lengths and evaluation metrics. 

In the ETTx dataset, the data for ETTm1 and ETTm2 are collected by sensors at a frequency of once every 15 minutes, while the data for ETTh1 and ETTh2 are collected once per hour. In Figure \ref{figure:ETT}, we can clearly observe that the ETTm1 dataset focuses primarily on short-term fluctuation features, while the ETTh1 dataset requires a more in-depth exploration of its complex periodic features. PatchTST performs remarkably well on the ETTm1 and ETTm2 datasets. However, in the ETTh1 and ETTh2 datasets, it falls behind WDformer. This phenomenon indicates that, although WDformer is not as effective as PatchTST in extracting short-term fluctuation features, it significantly outperforms PatchTST in extracting trend and periodic features. Moreover, due to the diverse characteristics and complex patterns of the ETT dataset, no model has yet achieved comprehensive superiority on the ETTh1 and ETTh2 datasets. 

In the ECL and Traffic datasets, WDformer demonstrates a more pronounced performance improvement compared to other models when it comes to prediction lengths of 336 and 720. This improvement is significantly greater than that observed at prediction lengths of 96 and 192. Additionally, in the Exchange dataset with a prediction length of 720, WDformer also achieves a substantial performance enhancement. We believe that the occurrence of this phenomenon is mainly attributed to the architecture of WDformer. First, the wavelet transform effectively removes high-frequency noise, allowing the model to focus more on low-frequency signals in long-term sequence prediction, thereby better capturing the long-term trends and core features of the data. Second, the differential attention mechanism further mitigates the interference of noise and enhances the model's ability to grasp trend and periodic patterns. These characteristics enable WDformer to perform exceptionally well in long-term prediction tasks.

\subsection{Ablation Study} % Study 是单数
To verify the effectiveness of the WDformer structure, we carefully designed a series of ablation experiments to systematically evaluate the impact of its key modules on model performance. In the ablation experiments, we established four different experimental configurations. The first group served as the baseline, without introducing either the wavelet transform or the differential attention mechanism. By comparing with the models that introduce these modules, we can more clearly see the specific impact of these modules on model performance. The second group introduced only the wavelet transform module. This means that the model relies solely on the raw time series data for prediction, without utilizing the multi-scale feature extraction capability provided by the wavelet transform. The third group introduced only the differential attention module. This means that the model only uses the traditional attention mechanism for prediction, without leveraging the differential attention mechanism to reduce noise and enhance the focus on key information. The fourth group is the complete WDformer model. WDformer is a time series forecasting model based on wavelet transform and differential attention mechanism proposed in this paper. It extracts multi-scale features of time series through wavelet transform and reduces noise by using a differential attention mechanism, thereby capturing more of the key information in the time series.

% As shown in Table \ref{tab:ablation}, we conducted experimental validation on the ECL and Traffic datasets for scenarios with prediction lengths of 96 and 192. When the prediction length is 96, both the wavelet transform and the differential attention mechanism demonstrate performance improvements on the ECL and Traffic datasets. Notably, when the two are applied together in the model, they further optimize the prediction accuracy, fully showcasing the powerful advantages of their synergistic effects. The model that only incorporates the differential attention mechanism achieves performance improvements across all prediction lengths and metrics on both datasets, compared with the baseline model. The wavelet transform version of the model exhibited a slight performance degradation only when the prediction length was 192 on the ECL dataset. On the Traffic dataset, when the prediction length is 96, in terms of the MSE metric, both the wavelet transform version and the differential attention version of the model show improvements compared with the baseline model. However, the result of WDformer is slightly worse than that of the differential attention version model. Overall, the differential attention module has a more significant impact on the overall improvement of the model, while the wavelet transform mechanism also plays an important synergistic role.

As shown in Table \ref{tab:ablation}, we conducted experimental validation on the ECL and Traffic datasets for forecast lengths of 96, 192, 336, and 720. On the ECL dataset, both the wavelet transform and the differential attention mechanism achieved performance improvements across all four forecast lengths, with the only exception being a slight accuracy drop in the wavelet transform at the 192 forecast length. This demonstrates the effectiveness of our two modules. Moreover, the improvement brought by the differential attention mechanism was greater than that of the wavelet transform mechanism, indicating that the differential attention mechanism is more adaptable to strongly periodic datasets like ECL. In particular, when both mechanisms were applied simultaneously, the prediction accuracy was further optimized, fully showcasing the powerful advantage of using them together. On the Traffic dataset, both the wavelet transform and the differential attention mechanism performed well at the 96 and 192 lengths. However, at the longer forecast lengths of 336 and 720, while the wavelet transform mechanism continued to perform well, the addition of the differential attention mechanism led to a decline in accuracy. At the 336 forecast length, the prediction accuracy of the differential attention mechanism was even lower than that of the baseline. This might be due to the fact that Traffic has more features, posing a challenge to the differential attention mechanism in long-range time forecasting. Overall, both the differential attention module and the wavelet transform mechanism contribute significantly to improving the model's overall performance, and in most cases, using them together leads to further accuracy gains.

\begin{table}[htbp]
\resizebox{\linewidth}{!}{
\begin{tabular}{@{}llllllllll@{}}
\toprule
\multicolumn{2}{l}{Design}                                               & \multicolumn{2}{l}{\begin{tabular}[c]{@{}l@{}}WDformer\\ - wave -DIFF\end{tabular}} & \multicolumn{2}{l}{\begin{tabular}[c]{@{}l@{}}WDformer\\ -DIFF\end{tabular}} & \multicolumn{2}{l}{\begin{tabular}[c]{@{}l@{}}WDformer\\ - wave\end{tabular}} & \multicolumn{2}{l}{WDformer} \\ \midrule
\multicolumn{2}{l|}{Metric}                                              & MSE                            & \multicolumn{1}{l|}{MAE}                           & MSE                        & \multicolumn{1}{l|}{MAE}                        & MSE                         & MAE                                             & MSE           & MAE          \\ \midrule
\multicolumn{1}{c|}{\multirow{4}{*}{ECL}}     & \multicolumn{1}{l|}{96}  & 0.148                          & \multicolumn{1}{l|}{0.240}                         & 0.146                      & \multicolumn{1}{l|}{0.239}                      & 0.145                       & \multicolumn{1}{l|}{0.237}                      & 0.144         & 0.237        \\
\multicolumn{1}{c|}{}                         & \multicolumn{1}{l|}{192} & 0.162                          & \multicolumn{1}{l|}{0.253}                         & 0.163                      & \multicolumn{1}{l|}{0.255}                      & 0.161                       & \multicolumn{1}{l|}{0.253}                      & 0.161         & 0.253        \\
\multicolumn{1}{c|}{}                         & \multicolumn{1}{l|}{336} & 0.178                          & \multicolumn{1}{l|}{0.269}                         & 0.174                      & \multicolumn{1}{l|}{0.268}                      & 0.173                       & \multicolumn{1}{l|}{0.267}                      & 0.173         & 0.267        \\
\multicolumn{1}{c|}{}                         & \multicolumn{1}{l|}{720} & 0.225                          & \multicolumn{1}{l|}{0.317}                         & 0.207                      & \multicolumn{1}{l|}{0.298}                      & 0.205                       & \multicolumn{1}{l|}{0.295}                      & 0.205         & 0.296        \\ \midrule
\multicolumn{1}{l|}{\multirow{4}{*}{Traffic}} & \multicolumn{1}{l|}{96}  & 0.395                          & \multicolumn{1}{l|}{0.268}                         & 0.394                      & \multicolumn{1}{l|}{0.265}                      & 0.392                       & \multicolumn{1}{l|}{0.268}                      & 0.391         & 0.265        \\
\multicolumn{1}{l|}{}                         & \multicolumn{1}{l|}{192} & 0.417                          & \multicolumn{1}{l|}{0.276}                         & 0.407                      & \multicolumn{1}{l|}{0.272}                      & 0.399                       & \multicolumn{1}{l|}{0.271}                      & 0.403         & 0.271        \\
\multicolumn{1}{l|}{}                         & \multicolumn{1}{l|}{336} & 0.433                          & \multicolumn{1}{l|}{0.283}                         & 0.416                      & \multicolumn{1}{l|}{0.277}                      & 0.450                       & \multicolumn{1}{l|}{0.293}                      & 0.423         & 0.280        \\
\multicolumn{1}{l|}{}                         & \multicolumn{1}{l|}{720} & 0.467                          & \multicolumn{1}{l|}{0.302}                         & 0.450                      & \multicolumn{1}{l|}{0.293}                      & 0.457                       & \multicolumn{1}{l|}{0.301}                      & 0.456         & 0.298        \\ \bottomrule
\end{tabular}
}
\caption{The results of ablation experiments. \textbf{"-wave"} indicates that the wavelet transform module is not introduced, while \textbf{"-DIFF"} indicates that the differential attention mechanism is not introduced.\@ \textbf{"-wave -DIFF"} indicates that neither of the two modules is introduced.}
\label{tab:ablation}
\end{table}

\begin{figure*}[htbp]
    \centering
    % 第一行子图
    \begin{subfigure}[b]{0.46\textwidth}
        \centering
    \includegraphics[height=5.6cm,width=1\textwidth]{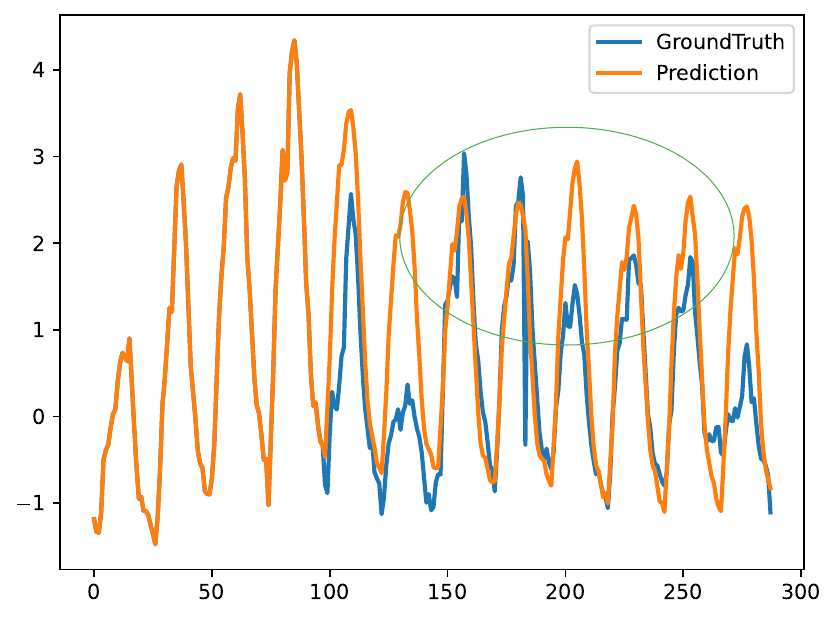}
        \caption{iTransformer}
        \label{ECL:sub1}
    \end{subfigure}
    \hfill % 添加间距
    \begin{subfigure}[b]{0.46\textwidth}
        \centering
    \includegraphics[height=5.6cm,width=1\textwidth]{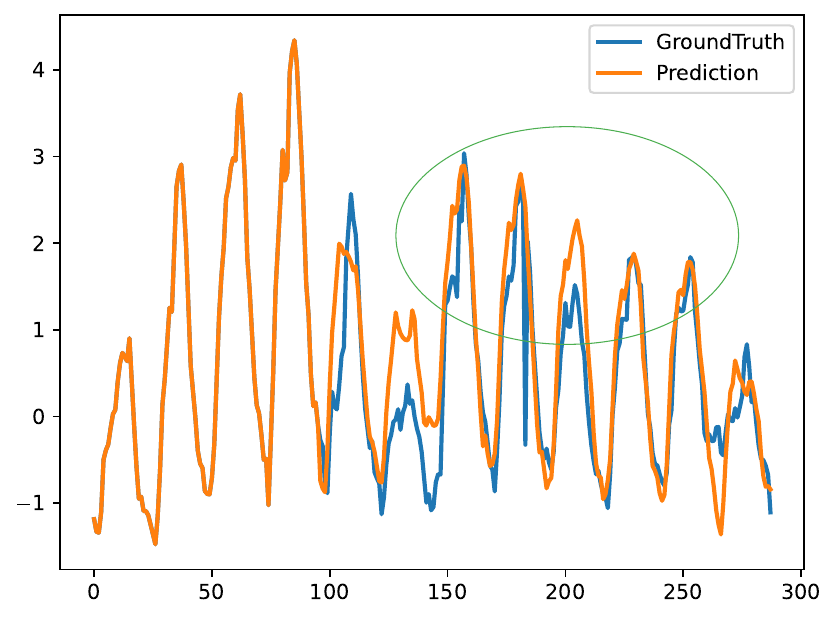}
    \caption{WDformer}
        \label{ECL:sub2}
    \end{subfigure}

    % 第二行子图
    \begin{subfigure}[b]{0.46\textwidth}
        \centering
        \includegraphics[height=5.6cm,width=1\textwidth]{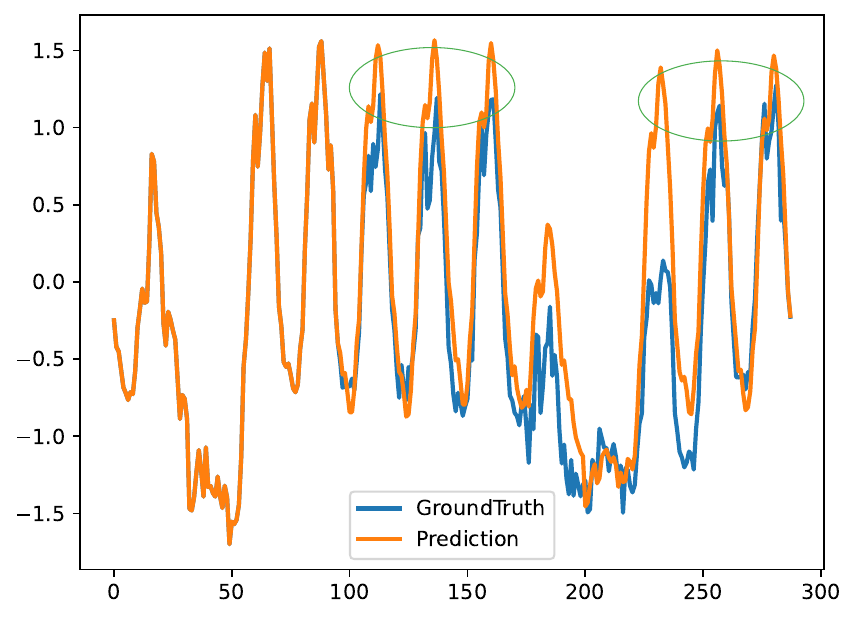}
         \caption{iTransformer}
        \label{ECL:sub3}
    \end{subfigure}
    \hfill % 添加间距
    \begin{subfigure}[b]{0.46\textwidth}
        \centering
    \includegraphics[height=5.6cm,width=1\textwidth]{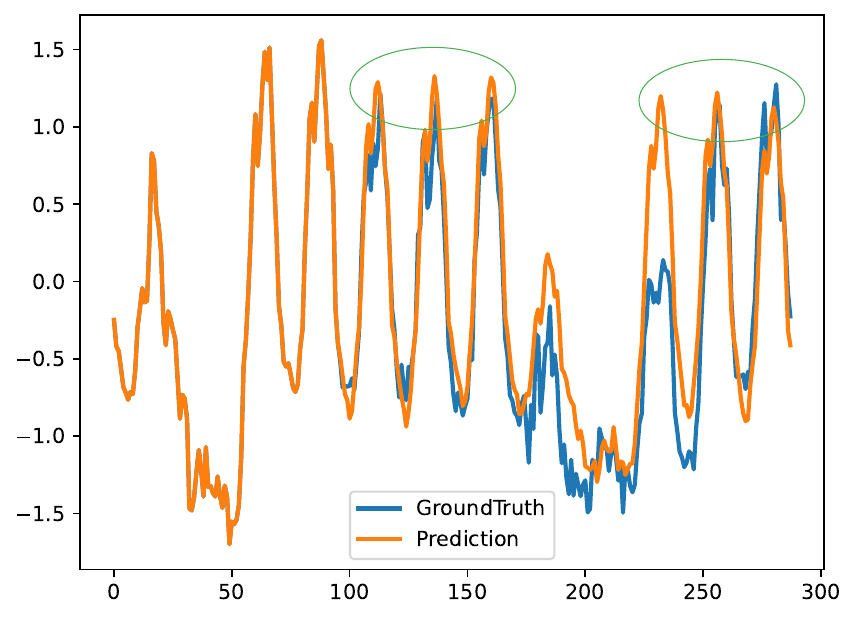}
    \caption{WDformer}
        \label{ECL:sub4}
    \end{subfigure}
    \caption{The forecasting results achieved by iTransformer and WDformer are based on the real data from the ECL dataset in the real world. The prediction results on the left are from iTransformer, while the prediction results on the right are from WDformer. The forecast results on the left and right sides are a comparative display for the same time period.}
    \label{figure:ECL}
\end{figure*}

\subsection{Visualization Analysis}

Taking iTransformer as the comparison benchmark, we conduct an in-depth analysis of the visualized images of the forecasting results. In Figure \ref{figure:ECL}, we show the visualized images of iTransformer and WDformer on the ECL dataset when the input length is 96 and the forecasting length is 192. It can be observed from the visualized images that the ECL dataset shows significant periodicity characteristics. iTransformer can accurately capture this kind of periodicity pattern during the forecasting process. However, compared with the actual amplitude changes of the peaks in the input sequence, there is still a certain deviation in its prediction for the peaks, and it fails to completely reproduce the significant changes of the peaks in the input sequence. The peaks of its forecasting results tend to be similar to those of the input sequence. By contrast, the amplitude of peak changes of WDformer is more intense, and it has a higher degree of fit with the real data, which is significantly better than the performance of iTransformer.

\begin{figure*}[htbp]
    \centering
    % 第一行子图
    \begin{subfigure}[b]{0.23\textwidth}
        \centering
        \includegraphics[height=3.3cm,width=1\textwidth]{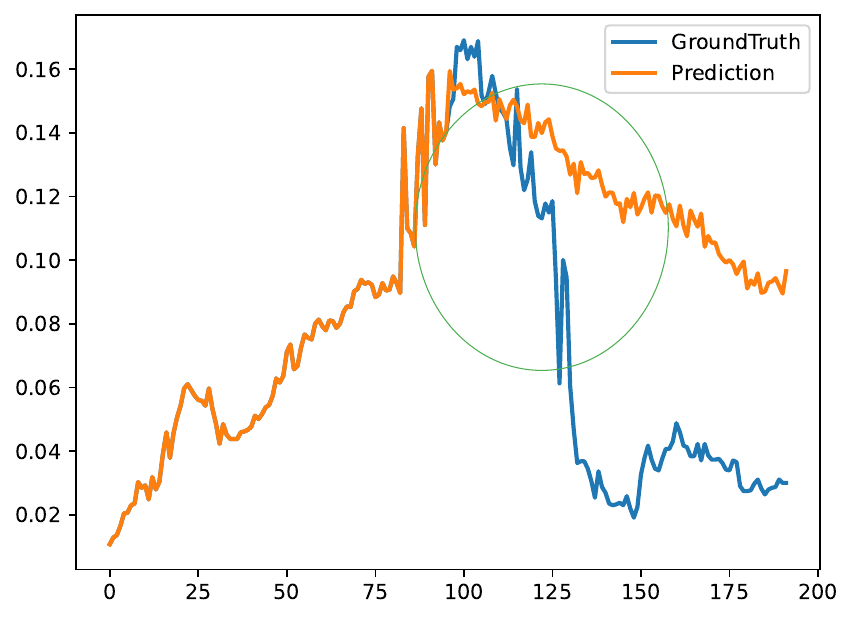}
        \caption{iTransformer}
        \label{Weather:sub1}
    \end{subfigure}
    \hfill % 添加间距
    \begin{subfigure}[b]{0.23\textwidth}
        \centering
        \includegraphics[height=3.3cm,width=1\textwidth]{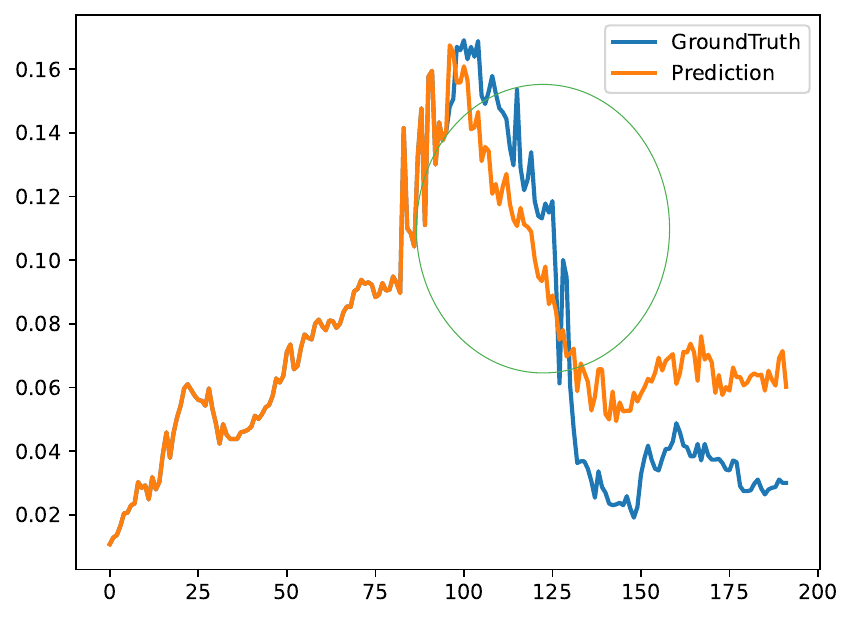}
        \caption{WDformer}
        \label{Weather:sub2}
    \end{subfigure}
    \hfill
    % 第二行子图
    \begin{subfigure}[b]{0.23\textwidth}
        \centering
        \includegraphics[height=3.3cm,width=1\textwidth]{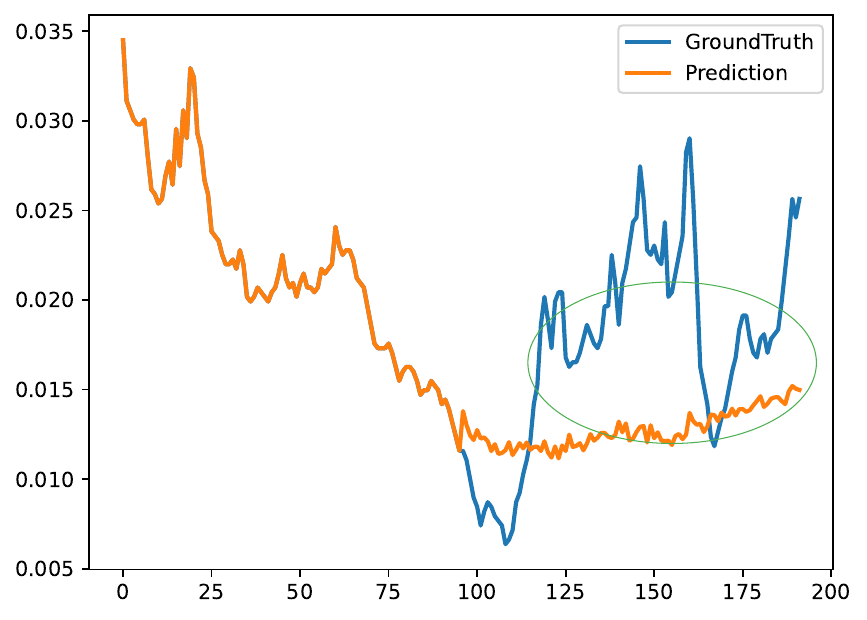}
        \caption{iTransformer}
        \label{Exchange:sub1}
    \end{subfigure}
    \hfill % 添加间距
    \begin{subfigure}[b]{0.23\textwidth}
        \centering
        \includegraphics[height=3.3cm,width=1\textwidth]{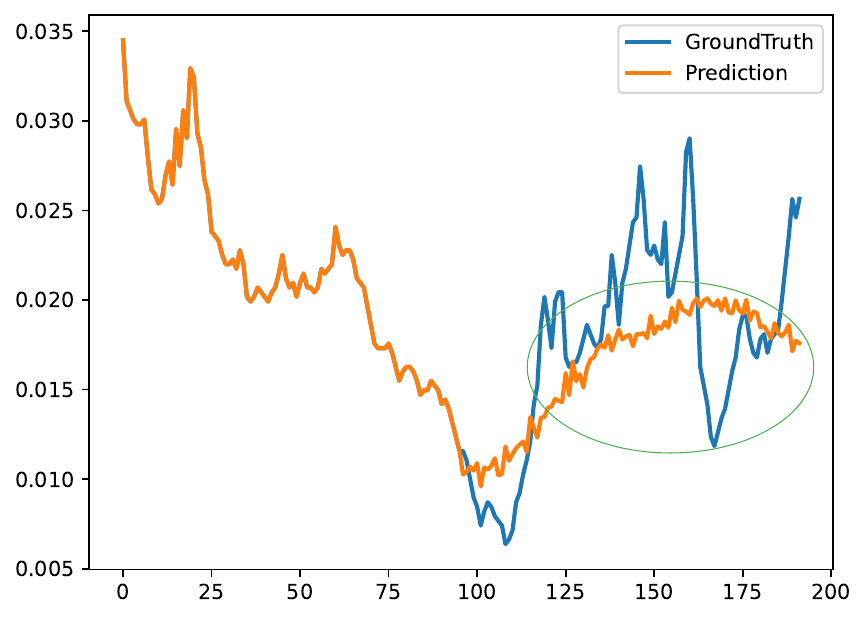}
        \caption{WDformer}
        \label{Exchange:sub2}
    \end{subfigure}
    \caption{The forecasting results achieved by iTransformer and WDformer are based on the real data from the Exchange dataset and Weather dataset in the real world. Subfigures \ref{Weather:sub1} and \ref{Weather:sub2} show the prediction results of the Weather dataset. Subfigures \ref{Exchange:sub1} and \ref{Exchange:sub2} show the prediction results of the Exchange dataset. The prediction results on the left are from iTransformer, while the prediction results on the right are from WDformer. Subfigures \ref{Weather:sub1} and \ref{Exchange:sub1} are from iTransformer, while Subfigures \ref{Weather:sub2} and \ref{Exchange:sub2} are from WDformer The forecast results are a comparative display for the same time period. }
    \label{figure:Exchange}
\end{figure*}

In Figure \ref{figure:Exchange}, we show the forecasting images of iTransformer and WDformer on the Exchange dataset and Weather dataset when the input length is 96 and the forecasting length is 96. The Exchange dataset includes the daily exchange rate data of eight countries from 1990 to 2016. The Weather dataset covers 21 key weather indicators, such as air temperature and humidity. The data in this dataset was recorded every 10 minutes throughout the year 2020. Different from the ECL power dataset with obvious periodicity characteristics, the Exchange dataset and Weather dataset have a weaker periodicity or a longer periodicity cycle. For such data, it is extremely important to accurately predict its trend. When forecasting the Exchange dataset and Weather dataset, iTransformer shows certain trend characteristics. When making predictions, iTransformer tends to hover around the height of the last time point of the input sequence and is more conservative in capturing the short-term fluctuations of the time series, relying more on the most recent historical information. By contrast, WDformer shows a bold forecasting strategy, with a greater amplitude of its peak changes and closer to the real data. 

Through the comparative analysis of the ECL, Weather, and Exchange dataset visualization results, it can be seen that iTransformer tends to adopt a more conservative strategy during the forecasting process. By contrast, WDformer is able to adopt a more aggressive strategy when facing significant changes in time series, and its forecasting results have a higher degree of fit with the real dataset. The original intention of iTransformer's design is to improve the forecasting performance of multivariate time series by flipping the structure of Transformer, which regards independent sequences as variable tokens and captures multivariate correlation through the attention mechanism. However, this design may have limitations in dealing with the complex patterns of time series, especially when facing significant changes in time series, as it lacks comprehensive consideration of time-domain and frequency-domain information. By contrast, the wavelet transform and differential attention mechanism of WDformer better capture the key information in time series, thereby showing higher flexibility and accuracy during forecasting. This mechanism enables the model to pay more attention to important features when facing significant changes in time series, so as to improve the closeness of the forecasting results.

\subsection{Complexity Analysis}
The time complexity of WDformer is mainly composed of the wavelet transform, embedding layer, differential attention mechanism, and feed-forward network. Among them, the time complexity of the wavelet transform is linear, $O\left ( K \right ) $, where $K$ is the length of the input time series. The time complexity of the embedding layer and the feed-forward network is $O\left ( Kd \right ) $, where $d$
represents the dimension of the embedding vector. The differential attention mechanism has a complexity of $O\left ( N^{2}d  \right ) $, where $N$ is the number of variables, because it needs to independently compute two attention matrices, making its complexity twice that of the traditional attention mechanism. WDformer is similar to iTransformer, both capturing the relationships between features through attention mechanisms. Since the time complexity of its attention mechanism is not affected by the length of the input time series, it has an advantage over other models when dealing with longer input sequences.

In terms of space complexity, WDformer, due to the introduction of the differential attention mechanism, has a storage requirement for its attention matrix that is twice that of the standard Transformer. However, apart from this aspect, the rest of WDformer's parameter scale is essentially on par with that of the standard Transformer.

We have verified the model's operational efficiency and memory usage on the ECL dataset. The dataset contains 25,635 records, which we divided into training, validation, and test sets in a 7:1:2 ratio. The device used for verification was an NVIDIA GeForce RTX 4090 with a memory capacity of 24 GB. We conducted a total of ten training epochs for the model, with each epoch maintaining an average training time of 33 seconds. During the training process, the GPU memory usage remained at 2.54 GB. The results presented above are derived from the average values of three independent experiments, ensuring the reliability and reproducibility of the data. For comparison, we conducted experiments on the iTransformer model with identical parameter settings. The iTransformer model achieved a stable average training time of 15 seconds per epoch. During the training process, the GPU memory usage remained at 1.72 GB.

\section{Conclusions}

This paper proposes a Transformer model based on wavelet transform and differential attention mechanism, aiming to improve the performance of time series forecasting. By integrating the multi-scale feature extraction capability of wavelet transform and the differential attention mechanism with dynamic weight allocation, the model can capture the key information in time series more accurately. Experimental results show that the model has achieved the SOTA experimental results on multiple real-world datasets, demonstrating its strong advantages in complex time series modeling.

\section{GenAI Disclosure section}
No GenAI tools were used in any part of this paper.

\bibliographystyle{ACM-Reference-Format}
\balance
\bibliography{sample-base}

% \appendix

\end{document}